\documentclass[journal]{IEEEtran}
\def\vec#1{\mathbf{#1}}
\usepackage{bm}
\usepackage{url}
\usepackage{amssymb}
\usepackage{amsmath}
\usepackage{amsfonts}
\usepackage{comment}
\usepackage[pdftex]{graphicx}

\ifCLASSINFOpdf
\else
\fi
\begin{document}
%
\title{Efficient Transfer Bayesian Optimization with Auxiliary Information}
%
%
%

\author{Tomoharu~Iwata,
  and~Takuma~Otsuka
\IEEEcompsocitemizethanks{\IEEEcompsocthanksitem T. Iwata and T. Otsuka are with NTT Communication Science Laboratories.}
\thanks{\copyright 2019 IEEE.  Personal use of this material is permitted.  Permission from IEEE must be obtained for all other uses, in any current or future media, including reprinting/republishing this material for advertising or promotional purposes, creating new collective works, for resale or redistribution to servers or lists, or reuse of any copyrighted component of this work in other works.}}

\maketitle

\begin{abstract}
We propose an efficient transfer Bayesian optimization method, which finds the maximum of an expensive-to-evaluate black-box function by using data on related optimization tasks. Our method uses auxiliary information that represents the task characteristics to effectively transfer knowledge for estimating a distribution over target functions. In particular, we use a Gaussian process, in which the mean and covariance functions are modeled with neural networks that simultaneously take both the auxiliary information and feature vectors as input. With a neural network mean function, we can estimate the target function even without evaluations. By using the neural network covariance function, we can extract nonlinear correlation among feature vectors that are shared across related tasks. Our Gaussian process-based formulation not only enables an analytic calculation of the posterior distribution but also swiftly adapts the target function to observations. Our method is also advantageous because the computational costs scale linearly with the number of source tasks. Through experiments using a synthetic dataset and datasets for finding the optimal pedestrian traffic regulations and optimal machine learning algorithms, we demonstrate that our method identifies the optimal points with fewer target function evaluations than existing methods.
\end{abstract}


%
\IEEEpeerreviewmaketitle

\section{Introduction}\label{sec:introduction}
%
%
%
%

\IEEEPARstart{B}{ayesian}
optimization (BO) is an approach for
the global optimization of noisy and black-box functions
that are expensive to evaluate~\cite{pelikan1999boa,brochu2010tutorial,shahriari2016taking}.
BO has outperformed other state of the art global optimization algorithms
on a number of benchmark functions~\cite{jones2001taxonomy},
and has been successfully used for a wide variety of applications,
such as the
automatic selection of machine learning algorithms~\cite{snoek2012practical,bergstra2013hyperopt,thornton2013auto,kotthoff2017auto,klein2017fast},
computer vision~\cite{denil2012learning},
probabilistic programs~\cite{rainforth2016bayesian},
sensor set selection~\cite{garnett2010bayesian},
chemical design~\cite{griffiths2017constrained},
material science~\cite{seko2015prediction,ueno2016combo},
and experimental particle physics~\cite{ilten2017event}.
BO uses a probabilistic model that is relatively inexpensive to evaluate
as a surrogate for expensive target functions.
Gaussian processes (GPs)~\cite{rasmussen2006gaussian}
are commonly used for surrogates to model a distribution over target functions.
The standard BO setting starts the optimization from scratch,
assuming no prior knowledge about a target function.
This lack of knowledge may sacrifice additional function evaluations for exploring the target function.

In some applications,
the target optimization task would be related to the tasks that have been given before.
An example is finding traffic regulations to reduce congestion
with different situations in an area.
When the number of pedestrians on streets is different,
the optimal traffic regulations are also different.
Nevertheless, the optimal traffic regulations in certain situations
can be optimal traffic regulations in different but related situations.
Another example 
is the selection of machine learning algorithms
with different datasets.
If an algorithm works well in a dataset,
it is also likely to
work well in similar datasets.
In these cases, 
the data in related optimization tasks often support
the optimization of a target function.

In this paper, we propose a transfer BO method that
can effectively and efficiently
transfer knowledge on source optimization tasks to the target task.
Our method uses auxiliary information that represents the task characteristics 
to transfer knowledge for estimating a distribution over target functions. 
In the example of the machine learning algorithm selection, 
the auxiliary information is such dataset's characteristics
as the sample and feature sizes.
For traffic regulation selections,
the observed number of people can be used as auxiliary information.

Our proposed method uses a Gaussian process 
whose mean and covariance functions are modeled with neural networks
that simultaneously take both the auxiliary information and feature vectors
as input.
By using the neural network mean function,
the GP aims at reliable estimations of the target function far from
the target observations
or even without any target observations.
With the neural network covariance function,
the GP can extract
nonlinear correlations between feature vectors
that are shared across source and target tasks,
which would help estimate the target function with fewer evaluations.
The mean and covariance functions define the prior distribution of the target function,
and the posterior distribution given target observations is calculated
in a closed form with the GP framework.
The nonparametric flexibility of the GP framework enables
us to rapidly capture the target function with fewer target observations.
The computational complexity for training the proposed model linearly increases
with the number of source tasks,
and that for calculating the posterior distribution
does not depend on the number of source tasks.
Therefore, the proposed method scales well to many source tasks.

Our technical contribution is that we develop a new method for Bayesian optimization with the following two advantages:
\begin{enumerate}
\item Our method can effectively transfer knowledge in other tasks using auxiliary information.
\item It can be trained efficiently with many tasks, where the computational cost for training scales linearly with the number of tasks.
\end{enumerate}
The remainder of this paper is organized as follows.
In Section~\ref{sec:related},
we briefly review related work.
In Section~\ref{sec:proposed},
we define our task and propose our method for a transfer BO
based on NGPs.
In Section~\ref{sec:experiments},
we demonstrate the effectiveness of our proposed method
using a synthetic dataset and two datasets for
finding optimal pedestrian traffic regulations
and optimal machine learning algorithms.
Finally, we present concluding remarks and a discussion of future work
in Section~\ref{sec:conclusion}.

\section{Related work}
\label{sec:related}

Many multi-task BO methods have been proposed, especially 
for hyperparameter optimization~\cite{swersky2013multi,snoek2015scalable,bardenet2013collaborative,poloczek2016warm,yogatama2014efficient}. 
For example, multi-task BOs based on multi-task GPs~\cite{swersky2013multi}
model functions in different tasks
by considering their correlation.
GPs can flexibly adjust the posterior distribution by their
nonparametric properties,
and work well with relatively few observations. 
This advantage is crucial for finding the optimum with fewer evaluations for BO.
Nevertheless, their computational complexity grows cubically
with the total number of observations. 
Therefore, it is computationally expensive to use data in many related tasks.
Multi-task BOs based on Bayesian neural networks~\cite{snoek2015scalable} 
scale well to many observations
and can approximate functions with high precision.
However, neural networks generally require many
observations to achieve good performance.
In addition, these methods do not use auxiliary information on tasks.
Although previous works~\cite{perrone2018scalable,wistuba2016two,schilling2015hyperparameter}
used auxiliary information,
they did not learn nonlinear interactions between the auxiliary information and black-box functions.
On the other hand, the proposed method learns them by incorporating neural networks
in both the mean and covariance functions with the GP framework.

Neural networks were previously incorporated into GPs~\cite{wilson2011gaussian,huang2015scalable,calandra2016manifold,wilson2016deep,wilson2016stochastic,iwata2017improving}.
For example, neural network covariance functions were
previously used~\cite{wilson2016deep},
as were neural network mean functions~\cite{iwata2017improving}.
However, these methods did not incorporate the auxiliary information
and were not designed for transfer learning or BO.
In addition, no previous work models both the mean and covariance functions 
with neural networks.
However, we model both the mean and covariance functions of GPs
with neural networks,
and their efficacy is demonstrated in our experiments.
Neural processes (NPs)~\cite{garnelo2018conditional}
model a distribution over functions based on neural networks.
With NPs, an encoder network embeds pairs of the features and function values
into a representation space,
and a decoder network estimates the mean and variance of the function value
given the embedded representation and a feature vector.
NPs, which are scalable, can use data in source tasks by neural networks.
However, since NPs are parametric models,
they are less flexible for adaptation to the given target observations
than GPs, which are nonparametric models.
In contrast, our GP exploits the nonparametric nature of swift adaptation
to the target observations,
even though the mean and covariance functions are modeled parametrically.

\section{Proposed method}
\label{sec:proposed}

In this section, we first give our problem setup for transfer BO in Section~\ref{sec:task}.
Then Section~\ref{sec:ngp} constructs our GP as a surrogate function model where the neural networks are trained with source tasks.
Section~\ref{sec:tbo} describes how the trained GPs are used to optimize a target task in the transfer BO framework.

\subsection{Task}
\label{sec:task}

Suppose that we are given data on $D$ source optimization tasks,
$\{\{\vec{x}_{dn},y_{dn}\}_{n=1}^{N_{d}},\vec{r}_{d}\}_{d=1}^{D}$,
where $\vec{x}_{dn}\in\mathbb{R}^{M}$ is the $n$th feature vector of task $d$,
$y_{dn}=f_{d}(\vec{x}_{dn})+\epsilon$ is its noisy scalar value of
black-box function $f_{d}(\cdot)$,
$\epsilon$ is the observation noise,
$N_{d}$ is the number of observations of task $d$,
$\vec{r}_{d}\in\mathbb{R}^{S}$ is the task descriptor of task $d$,
and $D$ is the number of source tasks.
Our goal is to find the maximum of unseen target function $\arg\max_{x}f_{d^{*}}(\vec{x})$
given target task descriptor $\vec{r}_{d^{*}}$.
We expect source functions $\{f_{d}(\cdot)\}_{d=1}^{D}$
to be informative for optimizing the target function.
For example, in the case of the hyperparameter optimization of machine learning algorithms
with many datasets,
feature vector $\vec{x}$ represents a hyperparameter configuration,
function value $y$ is its test accuracy,
and task descriptor $\vec{r}$ represents a dataset's properties,
such as sample and feature sizes.
Table~\ref{tab:notation} shows our notation.

\begin{table}
  \centering
  \caption{Notation.}
  \label{tab:notation}
 \begin{tabular}{ll}
  \hline
  Symbol & Description\\
  \hline
  $\vec{x}_{dn}$ & $n$th feature vector in task $d$\\
  $y_{dn}$ & $n$th scalar function value in task $d$\\
  $\vec{r}_{d}$ & task descriptor of task $d$\\
  $D$ & number of tasks\\
  $N_{d}$ & number of observations in task $d$\\
  $f_{d}(\cdot)$ & objective function of task $d$\\
  $m(\cdot;\bm{\xi})$ & neural network mean function\\
  & with parameter $\bm{\xi}$\\
  $g(\cdot;\bm{\psi})$ & neural network with parameter $\bm{\psi}$\\
  & used for a covariance function\\
  $k(\cdot;\bm{\theta})$ & covariance function with parameter $\bm{\theta}$\\
  \hline  
 \end{tabular}
  \end{table}
  
  \subsection{Gaussian Processes with Neural Mean and Covariance Functions}
  \label{sec:ngp}

\begin{figure}[t!]
 \centering
  \includegraphics[width=23em]{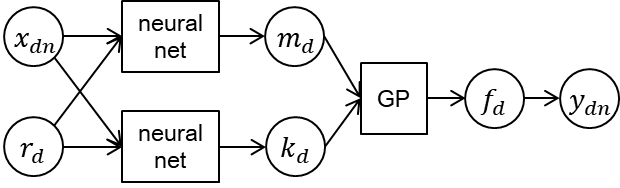}
  \caption{Our proposed model: Neural networks model task-specific mean function $m_{d}$ and covariance function $k_{d}$ by taking feature vector $\vec{x}_{dn}$ and task descriptor $\vec{r}_{d}$ as input. Gaussian process with mean and covariance functions generates task-specific function $f_{d}$, which outputs function value $y_{dn}$.}
  \label{fig:ngp}
 \end{figure}

We assume that the function of task $d$ is generated by the following
Gaussian process with task-specific mean and covariance functions:
\begin{align}
 f_{d}(\vec{x})\sim\mathcal{GP}\Bigl(m_{d}(\vec{x}),k_{d}(\vec{x},\vec{x}')\Bigr),
\end{align}
where $\mathcal{GP}(m,k)$ is the Gaussian process
with mean function $m$ and covariance function $k$,
$m_{d}(\vec{x})$ is the mean function of task $d$,
and $k_{d}(\vec{x},\vec{x}')$ is the covariance function of task $d$.
Although the task-specific mean and covariance functions help fit each task,
much data for each task are generally required for their estimation.
To overcome this difficulty,
we model the task-specific mean and covariance functions
with neural networks that are shared across different tasks:
\begin{align}
&m_{d}(\vec{x})=m(\vec{x},\vec{r}_{d};\bm{\xi}),\nonumber\\
&k_{d}(\vec{x},\vec{x}')= k\bigl(g(\vec{x},\vec{r}_{d};\bm{\psi}),g(\vec{x}',\vec{r}_{d};\bm{\psi});\bm{\theta}\bigr),
\end{align}
where 
$m(\cdot;\bm{\xi})$
is the mean function modeled by a neural network with parameter $\bm{\xi}$,
$k(\cdot,\cdot;\bm{\theta})$ is the covariance function with parameter $\bm{\theta}$,
$g(\cdot;\bm{\psi})$ is a neural network with parameter $\bm{\psi}$,
and the neural networks take 
both task descriptor $\vec{r}_{d}$ and feature vector $\vec{x}$ as input.
We call this the neural mean and covariance function Gaussian process (NGP).
The two neural networks provide task-specific mean and covariance functions
given task descriptor $\vec{r}_{d}$.
By incorporating the property of the tasks with the task descriptors,
NGPs can reduce the number of target observations necessary to fit the function.
When task descriptors are unavailable,
we simply omit $\vec{r}_{d}$ from the definition of $m(\cdot)$ and $g(\cdot)$:
$m(\vec{x};\bm{\xi})$, $g(\vec{x};\bm{\psi})$.
Parameters $\bm{\xi}, \bm{\psi}, \bm{\theta}$ are shared across all tasks,
making it unnecessary to train them for target tasks at test time 
since they are trained with source tasks in advance.
Figure~\ref{fig:ngp} illustrates our proposed NGP.

The mean function of GPs is usually set to zero since
GPs with zero mean functions and specific kernels
can approximate an arbitrary continuous function
given enough observations~\cite{micchelli2006universal}.
However, GPs with zero mean functions estimate zero
when the evaluation point is far from any other observations.
On the other hand, with non-zero mean function $m(\cdot;\bm{\xi})$,
NGPs can estimate the function even without any observations using task descriptors.
This transferred mean function is beneficial especially for Bayesian optimization settings,
where we estimate the function with fewer observations.
Feature vector $\vec{x}$ is transformed by neural network $g(\vec{x},\vec{r}_{d};\bm{\psi})$
before computing the covariance function by kernel $k(\cdot,\cdot;\bm{\theta})$.
Using this neural network attains flexible modeling of the correlation across feature vectors depending on task descriptor $\vec{r}_{d}$.

 When the Gaussian observation noise is assumed, $\epsilon\sim\mathcal{N}(0,\beta^{-1})$,
 where $\beta>0$ is the precision parameter, 
 by integrating out task-dependent functions $\{f_{d}(\cdot)\}_{d=1}^{D}$,
 the log likelihood of the NGP given data on source tasks is given:
\begin{align}
 L(\bm{\xi},\bm{\psi},\bm{\theta},\beta)
 = -\frac{1}{2}\sum_{d=1}^{D}\Bigl(N_{d}\log 2\pi+\log|\vec{K}_{d}+\beta\vec{I}|
 \nonumber\\
 +(\vec{y}_{d}-\vec{m}_{d})^{\top}(\vec{K}_{d}+\beta\vec{I})^{-1}(\vec{y}_{d}-\vec{m}_{d})
 \Bigr),
\end{align}
where
$\vec{y}_{d}=(y_{dn})_{n=1}^{N_{d}}$
is an $N_{d}$-dimensional column vector of the function values of task $d$,
$\vec{m}_{d}=(m(\vec{x}_{dn},\vec{r}_{d};\bm{\xi}))_{n=1}^{N_{d}}$
is the $N_{d}$-dimensional column vector of the mean function values, and
$\vec{K}_{d}$ is the $N_{d}\times N_{d}$ covariance matrix of task $d$, whose $(n,n')$ element is
given by $k(g(\vec{x}_{dn},\vec{r}_{d};\bm{\psi}),g(\vec{x}_{dn'},\vec{r}_{d};\bm{\psi});\bm{\theta}))$.
Parameters $\bm{\xi}$, $\bm{\psi}$, $\bm{\theta}$, $\beta$ are
optimized by maximizing the log likelihood
using stochastic gradient methods.
The computational complexity for training the NGPs is
$O(DN^{3})$ and it linearly scales with the number of source tasks,
where $D$ is the number of source tasks,
and $N$ is the number of observations in a source task.
When we use the linear kernel, its complexity is reduced to $O(DN)$.
In contrast, the computational complexity
of the existing multi-task GPs with nonlinear kernels is $O(D^{3}N^{3})$,
which is prohibitive for many tasks.

\subsection{Transfer Bayesian optimization}
  \label{sec:tbo}
  
Using an NGP trained with data on the source tasks,
we find the maximum of the target function
based on the Bayesian optimization framework.
For the target task, we iteratively
choose the next point to evaluate and update the NGP posterior with the obtained observation.

The next point to query is selected by finding the maximum of an acquisition function
for which we use the following
expected improvement criterion~\cite{mockus1978application,jones2001taxonomy},
 \begin{align}
  a(\vec{x})&=(\mu(\vec{x})-y^{\star})
  \Phi\left(\frac{\mu(\vec{x})-y^{\star}}{\sigma(\vec{x})}\right)
  +\sigma(\vec{x})\phi\left(\frac{\mu(\vec{x})-y^{\star}}{\sigma(\vec{x})}\right),
 \end{align}
 where $\phi(\cdot)$ and $\Phi(\cdot)$ are the probability density function and
 the cumulative density function of the standard normal.
 $y^{\star}$ is the maximum target function value obtained so far, and
 $\mu(\vec{x})$ and $\sigma(\vec{x})$ are the mean and standard deviation
 of the target function at $\vec{x}$.
 The expected improvement was efficient in the number of function evaluations
 required to find the global optimum of many multi-modal black-box functions~\cite{bull2011convergence}.
 Other acquisition functions can also be used in our framework, such as
 upper confidence bound~\cite{auer2002using,srinivas2010gaussian},
 entropy search~\cite{hennig2012entropy},
 predictive entropy search~\cite{hernandez2014predictive},
 and max-value entropy search~\cite{wang2017max}.

Assume that we observed $N_{d^{*}}$ points of target function $f_{d^{*}}$,
where the observed feature vectors are $\vec{X}^{*}=(\vec{x}_{d^{*}n})_{n=1}^{N_{d^{*}}}$,
and their function values are
$\vec{y}^{*}=(y_{d^{*}n})_{n=1}^{N_{d^{*}}}$.
With NGPs, given the target observations,
the posterior distribution of the target function is calculated
in a closed form by the following Gaussian distribution:
\begin{align}
 f_{d^{*}}(\vec{x})|\vec{X}^{*},\vec{y}^{*},\hat{\bm{\xi}}, \hat{\bm{\psi}}, \hat{\bm{\theta}}, \hat{\beta}
 \sim \mathcal{N}\left(\mu_{d^{*}}(\vec{x}),\sigma^{2}_{d^{*}}(\vec{x})\right),
\end{align}
\begin{align}
\mu_{d^{*}}(\vec{x})
 &=m(\vec{x},\vec{r}_{d^{*}};\hat{\bm{\xi}})
 \nonumber\\
&+\vec{k}_{*}^{\top}(\vec{K}_{*}+\hat{\beta}\vec{I})^{-1}(\vec{y}^{*}-m(\vec{x},\vec{r}_{d^{*}};\hat{\bm{\xi}})),
  \end{align}
\begin{align}
\sigma^{2}_{d^{*}}(\vec{x})=k_{\vec{x}}-
 \vec{k}_{*}^{\top}\vec{K}_{*}^{-1}
 \vec{k}_{*},
\end{align}
where
$k_{\vec{x}}=k(g(\vec{x},\vec{r}_{d^{*}};\hat{\bm{\psi}}),g(\vec{x},\vec{r}_{d^{*}};\hat{\bm{\psi}});\hat{\bm{\theta}}))$ is the scalar kernel value at $\vec{x}$,
$\vec{k}_{*}$ is the $N_{d^{*}}$-dimensional column vector of the kernel values
between $\vec{x}$ and $\vec{X}^{*}$,
$\vec{K}_{*}$ is the $N_{d^{*}}\times N_{d^{*}}$ kernel matrix
between the feature vectors in $\vec{X}^{*}$
given by
$k(g(\vec{x}_{d^{*}n},\vec{r}_{d^{*}};\hat{\bm{\psi}}),g(\vec{x}_{d^{*}n'},\vec{r}_{d^{*}};\hat{\bm{\psi}});\hat{\bm{\theta}}))$,
and $\hat{\bm{\xi}}$, $\hat{\bm{\psi}}$, $\hat{\bm{\theta}}$, and $\hat{\beta}$ are the parameters trained with the source data.

\section{Experiments}
\label{sec:experiments}

We investigated the efficacy of our method
using three datasets and compared it with existing approaches.
We examined which characteristics of NGPs,
such as task descriptors or mean and covariance networks,
contribute to the performance gain by ablation tests.

\subsection{Data}

We used the following three datasets:
Synthetic, Traffic, and Classifier data.
Table~\ref{tab:statistics} shows the statistics of the datasets.

  \begin{table}[t]
   \centering
  \caption{Statistics of datasets: number of tasks $D$,
  number of observations for each source task $N$,
  feature vector dimension $M$,
   and task descriptor vector dimension $S$.}
   \label{tab:statistics}
 \begin{tabular}{lrrrr}
  \hline
  Data & $D$ & $N$ & $M$ & $S$ \\
  \hline
  Synthetic & 100 & 500 & 1 & 1 \\
  Traffic & 100 & 1,125 & 8 & 3 \\
  Classifier & 80 & 229 & 23 & 11 \\
  \hline  
 \end{tabular}
  \end{table}

\begin{figure}[t!]
 \centering
  \includegraphics[width=26em]{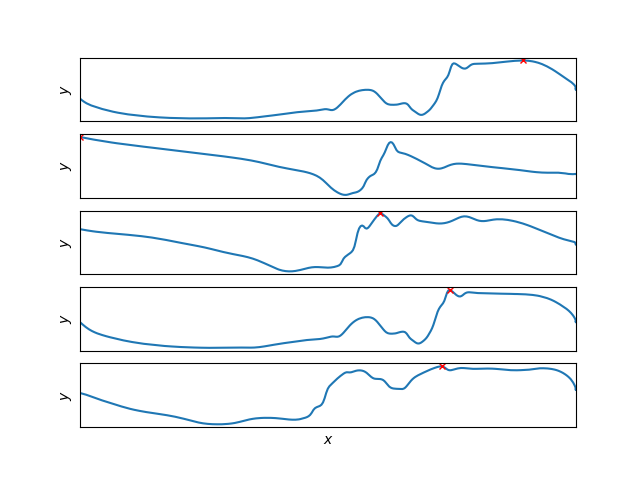}
  \\
  (a) Synthetic data
  \\
  \includegraphics[width=26em]{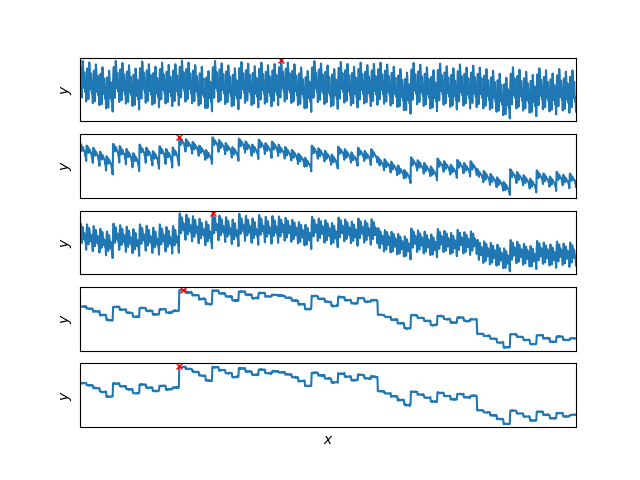}
  \\
  (b) Traffic data
  \includegraphics[width=26em]{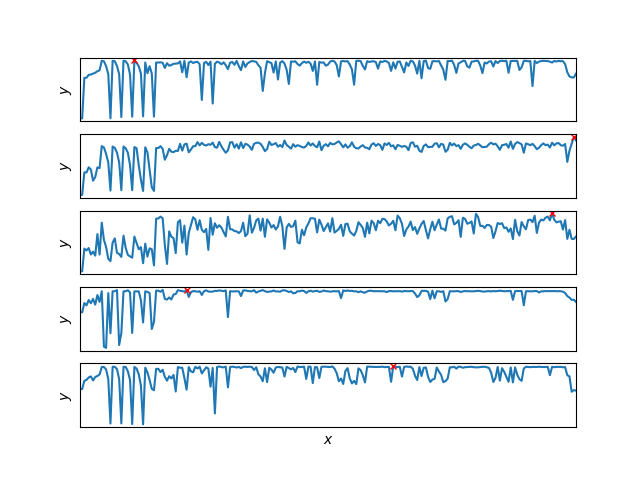}
  \\
  (c) Classifier data
  \caption{Examples of datasets:
  Horizontal axis is feature value $\vec{x}$ in (a), and 
  it is the feature vector index in (b) and (c).
  Vertical axis is function value $y$,
  and each line represents function value for each task.
  Red `$\times$' mark indicates maximum point in task.}
  \label{fig:data}
\end{figure}

 \begin{figure}[t!]
  \centering
  \includegraphics[width=25em]{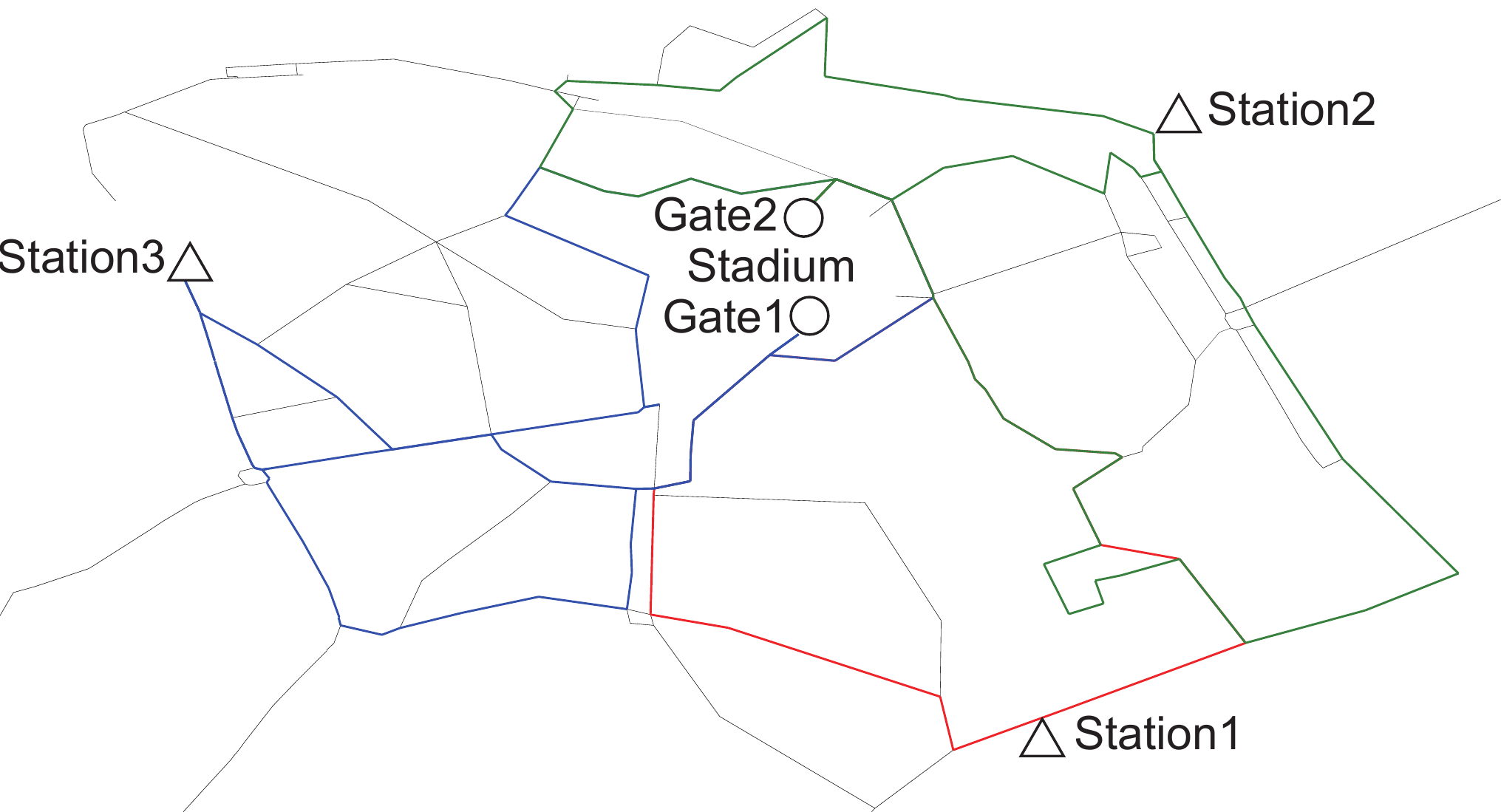}
  \caption{Map around National Stadium in Tokyo for pedestrian simulations with Traffic data. Three train stations (marked with `$\triangle$') are route origins, and two stadium gates (marked with `$\bigcirc$') are route destinations. Black line is a street, and colored lines are routes, where color indicates origin-destination pair (OD) index: 1st: red, 2nd: green, 3rd: blue.}
  \label{fig:map}
\end{figure}  

Synthetic data were artificially synthesized using GPs and neural networks.
First, we generated 500 evenly spaced scalar values from -5 to 5.
Second, one-dimensional feature vectors $\vec{x}$ were generated from these 500 values
using four-layer feed-forward neural networks with 32 hidden units.
Third, for each task,
one-dimensional task descriptor vector $\vec{r}$ was generated
from the standard normal distribution.
Fourth, the task descriptor and the evenly spaced scalar values were concatenated.
Fifth, we constructed two neural networks for mean function $m(\cdot)$
and covariance function $g(\cdot)$,
which were four-layer feed-forward neural networks with 32 hidden units and one output unit.
The parameters of the neural networks were determined uniform randomly.
Sixth, the concatenated vector was mapped to scalar value $y$ by an NGP
with the RBF kernel and neural networks constructed at the fifth step.
The numbers of source, validation, and target tasks were $D=100$, 20, and 20, respectively.
Figure~\ref{fig:data}(a) shows examples of the five tasks
on the Synthetic data.

Traffic data aims to find the optimal traffic regulation for pedestrians,
who were simulated using a multi-agent-based crowd simulator
using the streets near National Stadium in Tokyo shown in Figure~\ref{fig:map}.
The pedestrians walked from train stations to the stadium,
where each individual was assigned to one of three pairs
of origin and destination (OD):
from a certain station to an entrance gate of the stadium.
The first OD was Station1 and Gate1,
the second OD was Station2 and Gate2,
and the third OD was Station3 and Gate1.
A traffic regulation controlled the proportion of routes selected for each OD.
There were two routes for the first OD
and three routes for the second and third ODs.
Thus, a regulation consisted of three proportions,
such as 0:1, 0.5:0.5:0, and 0:0.25:0.75 for each OD pairs.
Feature vector $\vec{x}$ was $M=8$ dimensions,
which correspond to the concatenated proportions,
and
the combination of such proportion amounts to 1,125 regulations.
Function value $y$ was the rate of the pedestrians who arrived at their destination
before the one-hour simulation ended.
In this simulation, arrival rate $y$
may decrease with some regulations when the amount of traffic exceeds route capacities.
We generated data on 152 tasks with different numbers of pedestrians for ODs.
For example, a task was simulation results with
15,000, 5,000, and 20,000 pedestrians for each of the three ODs.
Task descriptor $\vec{r}$ represented the population of each OD
with $S=3$ dimensions.
We randomly generated 100 splits of source, validation, and target tasks,
whose sizes were $D=100$, 22, and 30.
and the results were averaged over 100 splits.
Figure~\ref{fig:data}(b) shows examples of the five tasks on the Traffic data.

Classifier data dealt with optimization tasks
to find the best combination of a classifier and its hyperparameters
for a given dataset.
We used the following six classifiers in scikit-learn~\cite{pedregosa2011scikit}:
k-nearest neighbor method,
support vector machines,
Gaussian processes,
random forests,
neural networks and Adaboost.
Feature vector $\vec{x}$ represented a classifier with its hyperparameters to be used,
where the classifier and categorical hyperparameters were encoded by one-hot vectors,
and these one-hot vectors and real-valued hyperparameters were concatenated.
In the feature vector, the hyperparameters of the unused classifiers were set to zero.
The dimension of the feature vector was $M=23$.
We used the following nine datasets
for the binary classification~\cite{chang2011libsvm}\footnote{The datasets were obtained from \url{https://www.csie.ntu.edu.tw/~cjlin/libsvmtools/datasets/binary.html}}:
Australian, Breast-cancer, Diabetes, German-numer,
Heart, Ionosphere, Liver-disorders, Sonar, and Splice.
For each original dataset,
we modified the feature and training data sizes to generate related source tasks
with different difficulty as a classification problem.
In particular,
we randomly selected features to use with probabilities 1.0, 0.7, 0.4
and randomly selected training data to use with probabilities 0.8, 0.6, 0.4, and 0.2.
We constructed 108 datasets: nine original datasets $\times$ three feature sizes $\times$ four training data sizes.
For each dataset, we calculated
the AUC (area under the receiver operating characteristic curve) on the test data
with 229 combinations of a classifier and hyperparameters
and used the AUC for function value $y$.
The task descriptor was constructed by concatenating the training data size, the feature size, and a one-hot vector that specified the original nine datasets,
which means that its total dimensions were $S=11$.
We generated 100 splits of source, validation, and target tasks,
whose sizes were $D=80$, 13, and 15,
and averaged the results over 100 splits.
Figure~\ref{fig:data}(c) shows examples of the five tasks on the Classifier data.

\subsection{Comparing methods}

We evaluated the following four versions of our proposed NGP based transfer BO:
NGP-RMK, NGP-RM, NGP-RK, and NGP-MK.
`R' indicates that it uses task descriptor $\vec{r}$.
`M' indicates that it uses neural network mean function $m(\cdot)$, and
the method without `M' (NGP-RK) used the zero mean function.
`K' indicates that it uses the neural network covariance function $g(\cdot)$,
and the method without `K' (NGP-RM)
directly applies the RBF kernel without neural networks.
For the mean neural networks, we used four-layer neural networks with 32 hidden units.
For the neural networks for covariance functions,
we used three-layer neural networks with 32 hidden and output units.
When the task descriptor was used,
we concatenated feature vector $\vec{x}$ and task descriptor $\vec{r}$,
and used the concatenated vector
for the input of the neural networks for mean and covariance functions.
We used the RBF kernel, and the batch size was 32.
The validation tasks were used for early stopping.
We optimized the neural network parameters and kernel parameters
using ADAM~\cite{Adam} with learning rate $10^{-2}$.

We compared our proposed method with the following six methods: Gaussian process (GP), transfer Gaussian process (TGP),
neural process (NP), neural network (NN), neural network with task descriptor $\vec{r}$ (NN-R), and Random.

GP denotes a standard BO method that uses the zero-mean Gaussian process with the Mat\'ern kernel.
The GP does not use data on the source tasks.
The TGP is a BO using the Gaussian process with the RBF kernel,
where the kernel parameters were trained using data on the source tasks.
The TGP is a simple form of a transfer BO method.

NP is a BO using a conditional neural process~\cite{garnelo2018conditional},
which is a stochastic model of functions based on
encoder and decoder neural networks.
The encoder 
transforms the set of feature and function value pairs of a task into a task-specific latent vector.
The decoder estimates the mean and variance of the function value
given the latent and feature vectors.
We used 32-dimensional latent vectors and three-layer neural networks with 32 hidden units
for the encoder and decoder.

NGP-RMK, NGP-RM, NGP-RK, NGP-MK, GP, TGP, and NP are all BO-based methods
with different methods for modeling objective functions.
For all the BO methods above, 
we used the expected improvement criterion for the acquisition function.

NN and NN-R are supervised neural network-based methods.
With NN, a four-layer neural network with 32 hidden units
was trained with feature vector $\vec{x}$ and function value $y$ pairs in the source tasks.
Then we selected the next point to query by finding
the maximum function value of the trained neural network.
NN-R uses task descriptor $\vec{r}$ as the input of the neural network of NN
by concatenation with feature vector $\vec{x}$.
NN and NN-R do not use the data on the target tasks.
Random selects the next point randomly.

\subsection{Results}

 \begin{table}[t!]
  \centering
  \caption{Average number of target function evaluations required to find maximum point and its standard error. Bold indicates best performing method, and star $\star$ indicates that it is not significantly different from best performing method based on a paired t-test with at 5\% level.}
  \label{tab:result}
 \begin{tabular}{lrrr}
   \hline
   & Synthetic & Traffic & Classifier \\
   \hline
  NGP-RMK & {\bf 9.00$\pm$1.85}$^{\star}$ &
	  {\bf 18.16$\pm$0.81}$^{\star}$
      & {\bf 60.40$\pm$1.58}$^{\star}$ \\ 
  NGP-RM &  19.50$\pm$3.78 &22.53$\pm$0.90 & 61.19$\pm$1.63$^{\star}$ \\  
  NGP-RK &  25.65$\pm$2.75 &23.38$\pm$0.82 & 62.87$\pm$1.52$^{\star}$ \\  
  NGP-MK & 25.85$\pm$2.40 & 19.82$\pm$0.66 & 61.32$\pm$1.64$^{\star}$ \\  
  GP & 71.05$\pm$30.04 & 42.30$\pm$1.00 & 78.59$\pm$1.85 \\  
  TGP & 27.95$\pm$3.39 & 31.16$\pm$0.85 & 88.00$\pm$1.56 \\  
  NP & 147.95$\pm$18.42 & 162.37$\pm$3.61 & 76.92$\pm$1.87 \\ 
  NN & 192.40$\pm$19.12 & 172.57$\pm$3.99 & 83.95$\pm$1.83 \\  
  NN-R & 66.45$\pm$12.28 & 35.41$\pm$1.20 & 70.05$\pm$1.80 \\  
  Random & 333.40$\pm$30.92 & 565.52$\pm$5.95 & 107.79$\pm$1.77\\ 
 \hline
 \end{tabular}
 \end{table}
 
 \begin{figure}[t!]
 \centering
\includegraphics[width=23em]{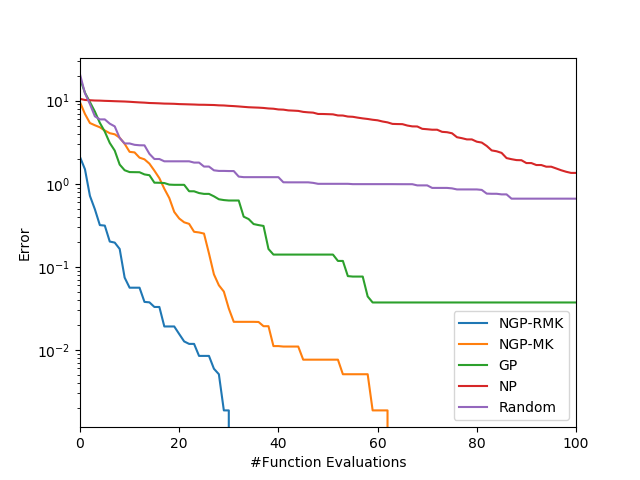}
\\
(a) Synthetic data
\\ 
\includegraphics[width=23em]{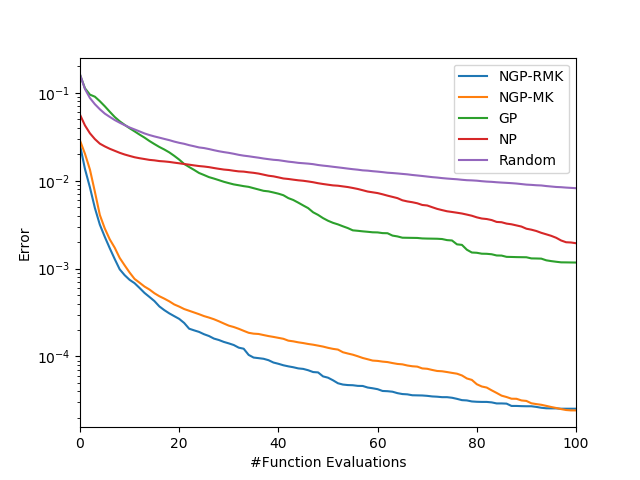}
\\
(b) Traffic data
\\
\includegraphics[width=23em]{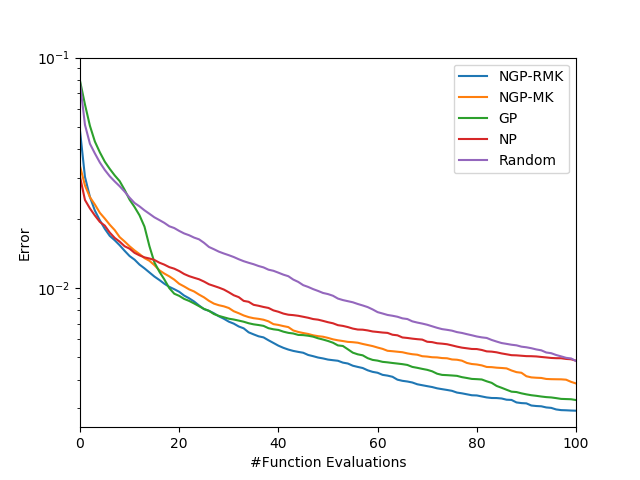}
\\
(c) Classifier data
\caption{Difference between true maximum value and
maximum value obtained so far with different numbers of function evaluations.}  
\label{fig:diff}
\end{figure}

 \begin{figure}[t!]
 \centering
  \includegraphics[width=27em,height=8.5em]{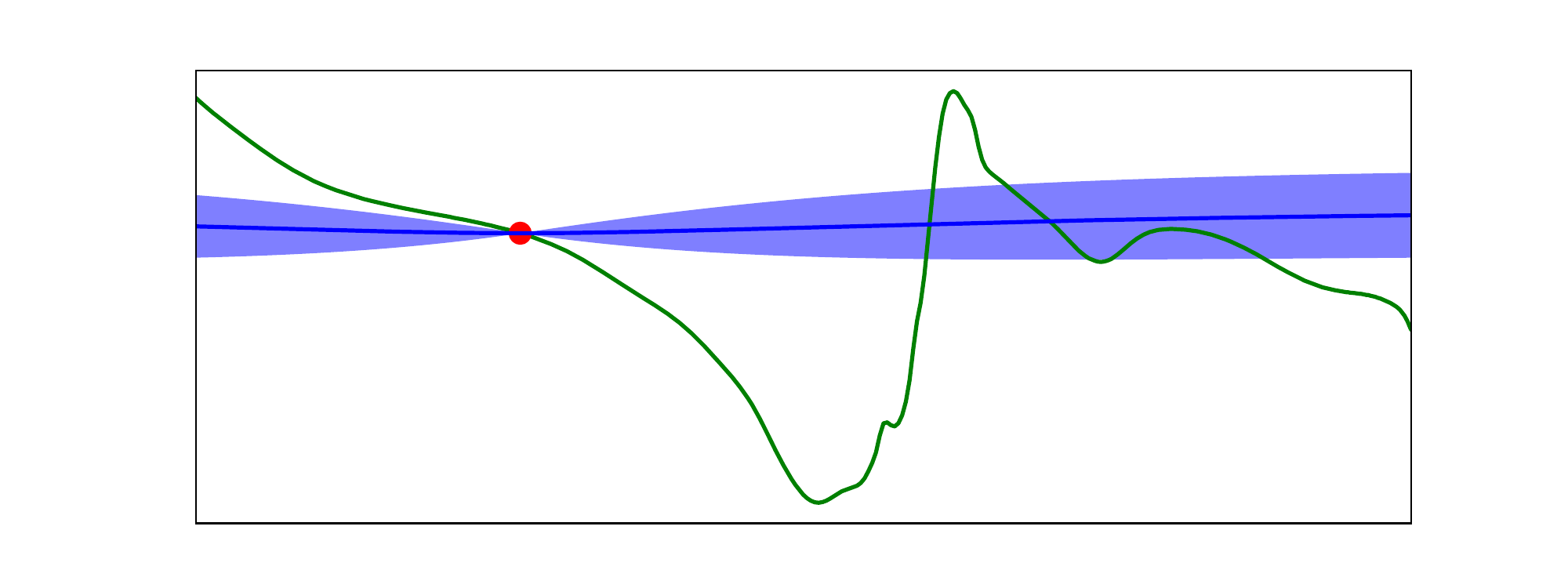}
  \\
  \#function evaluations: 1
  \\
  \includegraphics[width=27em,height=8.5em]{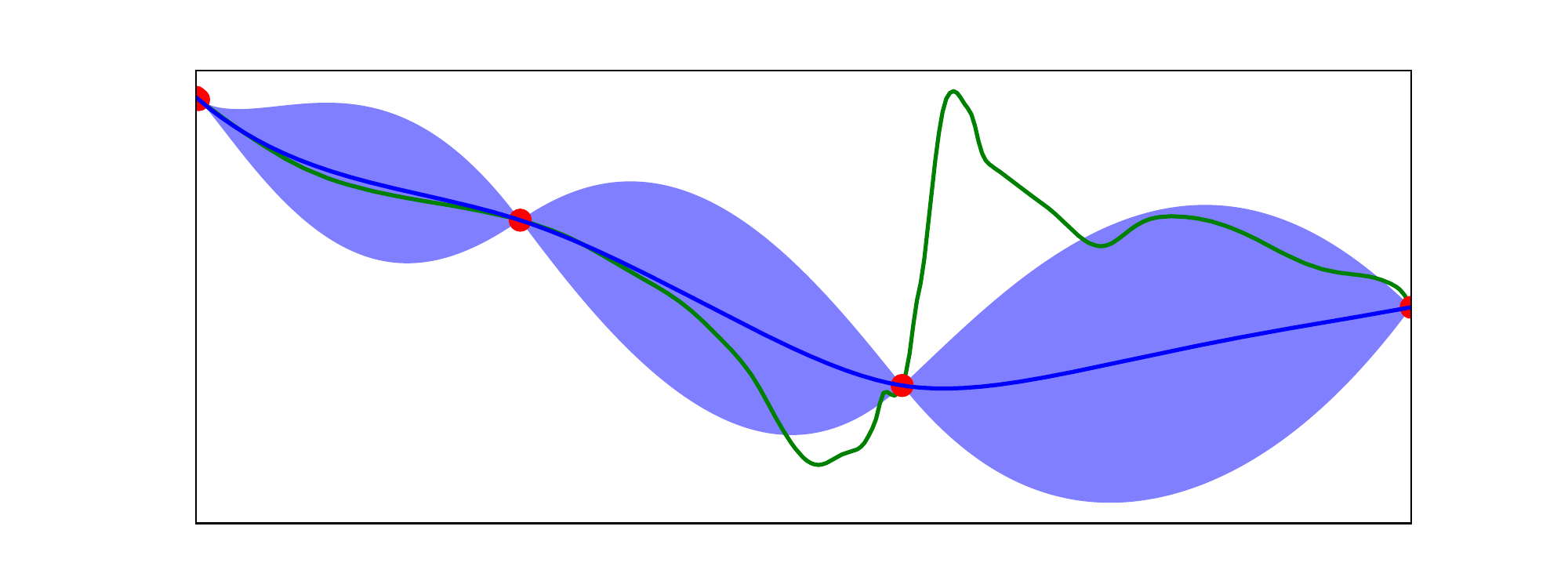}
  \\
  \#function evaluations: 5
  \\
  \includegraphics[width=27em,height=8.5em]{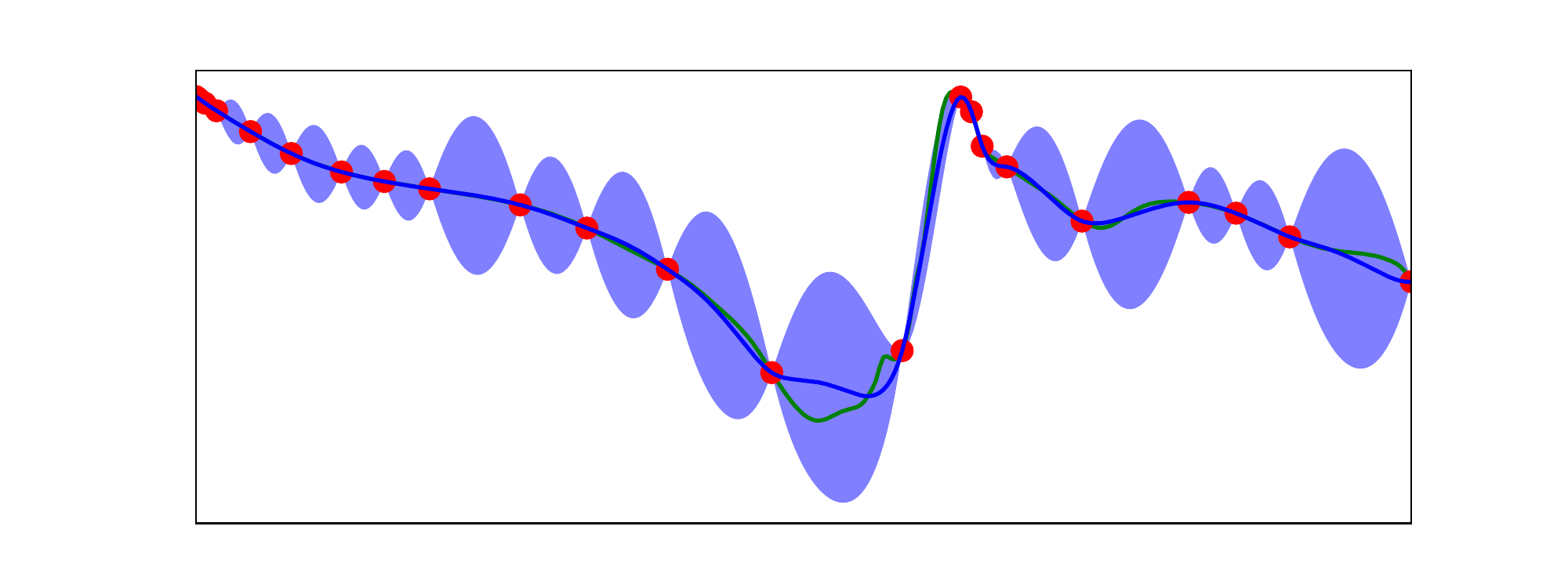}
  \\
  \#function evaluations: 23
  \\
  (a) GP
  \\
  \includegraphics[width=27em,height=8.5em]{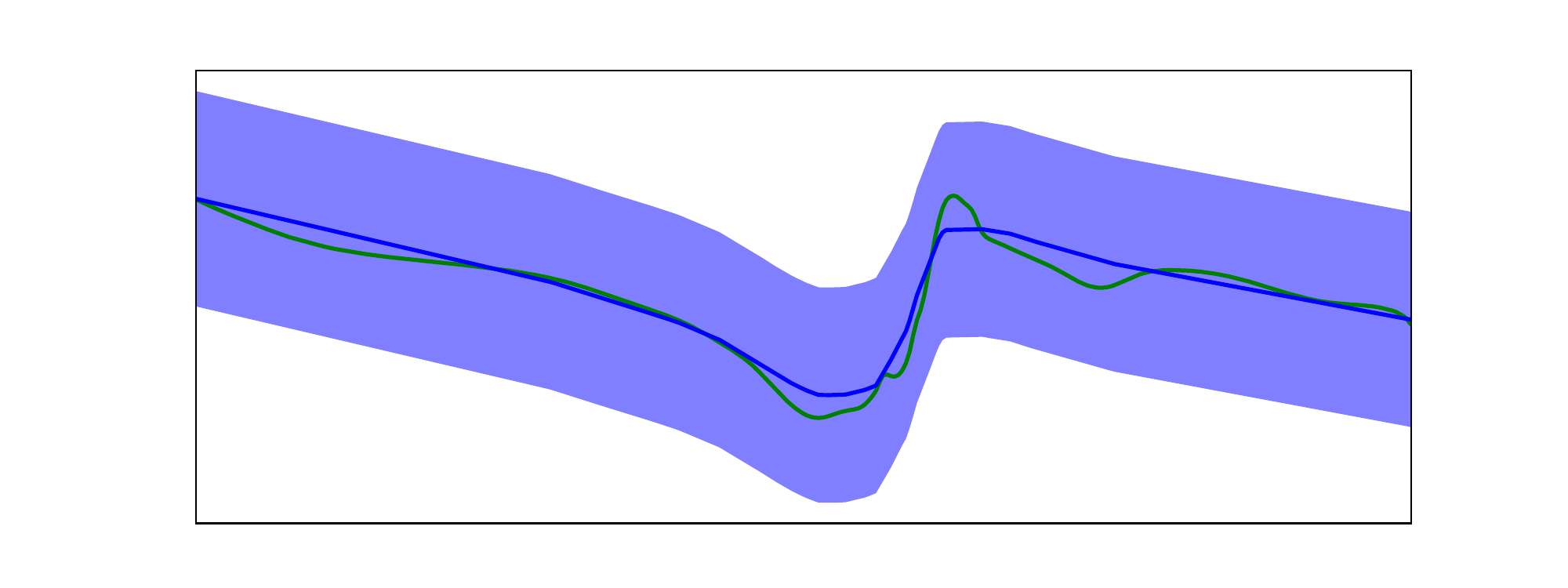}
  \\
  \#function evaluations: 0
  \\
  \includegraphics[width=27em,height=8.5em]{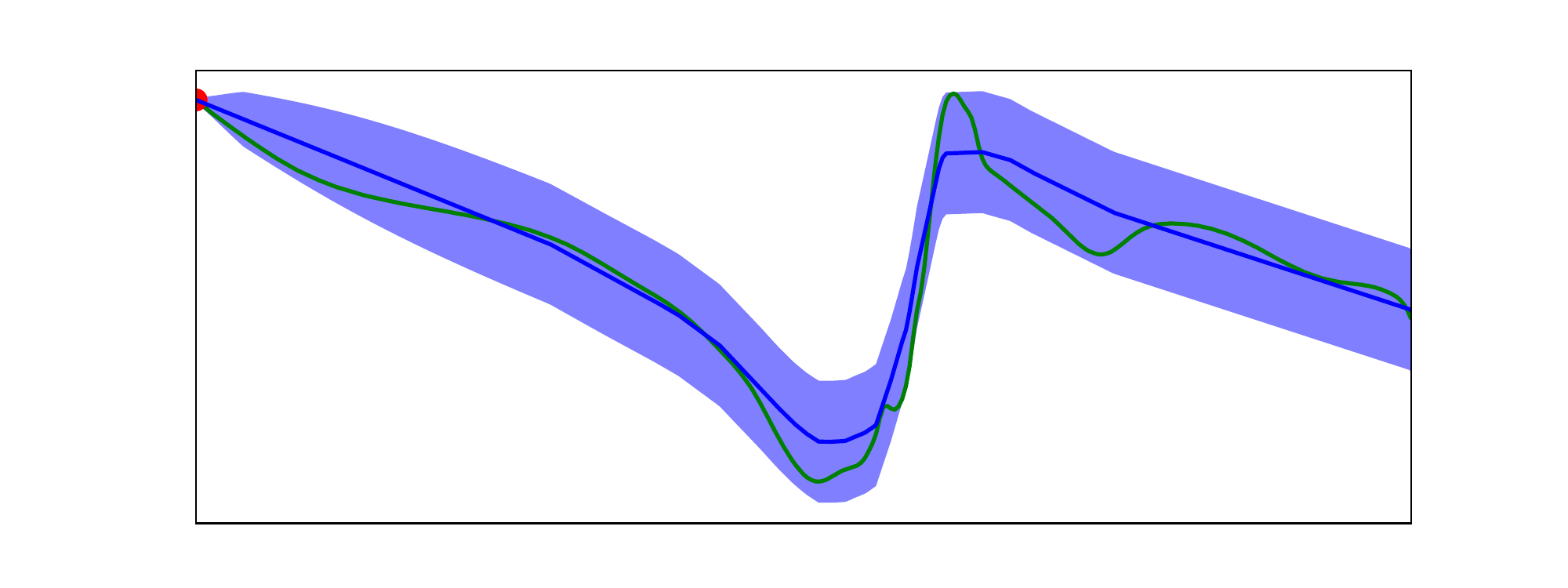}
  \\
  \#function evaluations: 1
  \\
  \includegraphics[width=27em,height=8.5em]{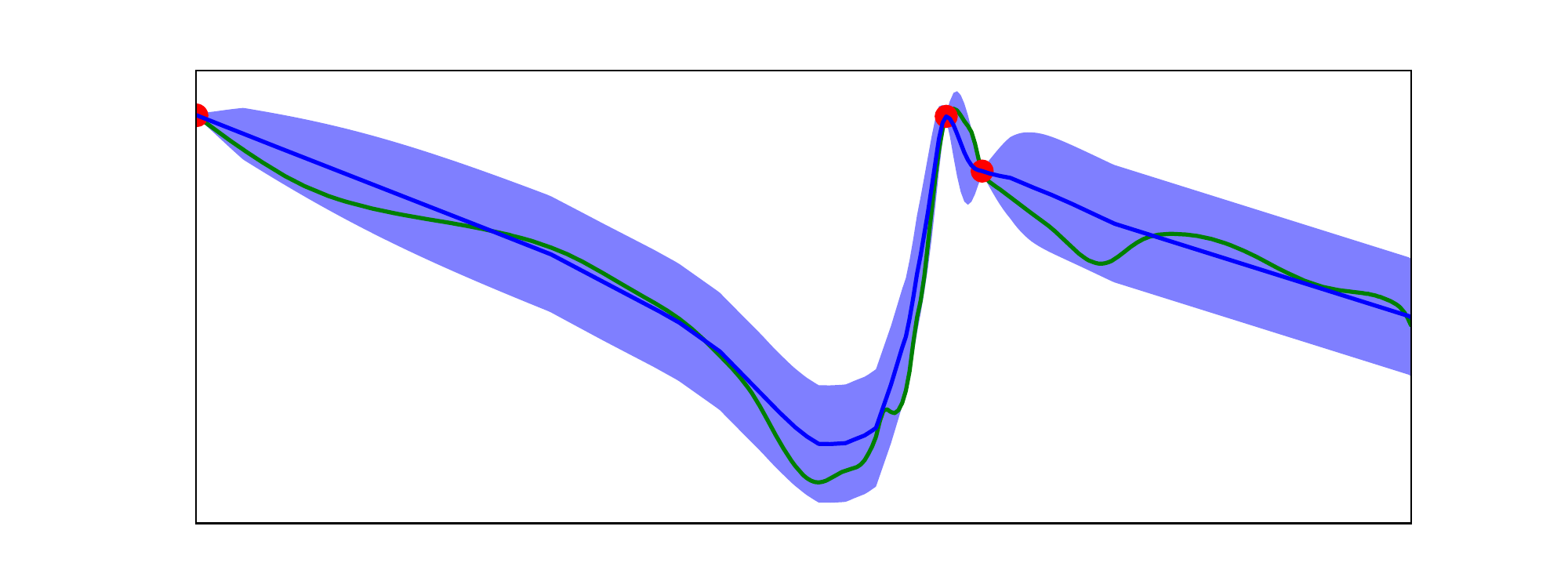}
  \\
  \#function evaluations: 4
  \\
  (b) NGP-RMK
  \caption{Estimated target functions on Synthetic data with different numbers of target function evaluations by GP (a) and NGP-RMK (b). Horizontal axis is feature value $\vec{x}$, and vertical axis is function value $y$. Green line is true function, blue line is estimated function, blue area is 95\% confidence interval, and red points indicate evaluated points.}
  \label{fig:function}
   \end{figure}

Table~\ref{tab:result} shows the average number of target function evaluations
required to find the maximum point.
The NGP-RMK achieved the best performance on all three datasets,
indicating that
the neural network mean and covariance functions as well as incorporating the task descriptors
are beneficial for the transfer BO method.
GP needed more function evaluations than NGP since
GP could not use the information in the source tasks.
Although TGP learned the kernel parameters using the data in the source tasks,
the kernel parameters lacked rich expressive power
to model functions in related tasks.
On the other hand, NGP effectively modeled the task characteristics
using neural networks.
NP performance was worse than NGP
because
NP pretrained the encoder and decoder neural networks using the source tasks.
It was difficult to flexibly fit the obtained data on unseen target tasks.
In contrast, NGP 
adaptively estimated the posterior distribution with the target data using the GP framework.
NN and NN-R used the data in the source tasks,
but they could not use them in the target tasks.
Therefore, they failed to model the target tasks that were different from the source tasks.

Figure~\ref{fig:diff}
shows the differences between the true maximum value and
the maximum value obtained so far with different numbers of function evaluations.
NGP-RMK minimized the error with fewer function evaluations.
When there were no function evaluations,
NGP-RMK, NGP-MK, and NP achieved low error since
they learned the distribution of the target functions using the source tasks.
Especially, NGP-RMK achieved the lowest error without function evaluations
with the Synthetic and Traffic data since NGP-RMK used the task descriptor.
Since GP could not use the information in the source tasks,
its error in the early stage
was almost the same with Random.
NP did not effectively decrease the error compared with the other methods since
it used pretrained neural networks for modeling the target functions.

Figure~\ref{fig:function} shows
the estimated target functions using the Synthetic data
with different numbers of target function evaluations by GP (a) and NGP-RMK (b).
Although GP fit the true function near the observed points,
it did not give good estimations far from the observed points.
As a result, it required 24 function evaluations to find the maximum in this example.
On the other hand, NGP-RMK gave a reasonable estimation without any function evaluations.
It also reduced the variance near the observed points,
and quickly fit the observed target function value.
With this flexibility,
even when the prior distribution of the functions is far from the target function,
NGP-RMK can fit the target function by observing the target function values.
NGP-RMK found the maximum point using only five function evaluations in this example.

  \section{Conclusion}
 \label{sec:conclusion}

 We proposed a transfer Bayesian optimization method that
 identifies the maximum of a target black-box function by utilizing
 data on related optimization tasks.
 With our proposed method, Gaussian processes with neural mean and covariance functions
 are used for modeling distributions over functions.
 By training the neural mean and covariance functions using the data on related tasks,
 we found the maximum with fewer function evaluations.
 For future work, 
 we plan to
 extend NGPs to structured features, such as sequences, images and graphs
 using recurrent, convolutional and graph convolutional neural networks~\cite{mikolov2010recurrent,krizhevsky2012imagenet,kipf2016semi}.
 Since NGPs learn covariance structure with neural networks,
 they should be robust to high dimensional feature space~\cite{wang2017batched,li2017high,li2016high,pmlr-v70-rana17a,wang2016bayesian}.
 We also plan to leverage techniques for BO and GPs into our framework
 to improve the scalability with the number of observations in a task
 using sparse GPs~\cite{hensman2013gaussian} and 
 to handle constraints~\cite{gardner2014bayesian,hernandez2016general}
 and batched queries~\cite{gonzalez2016batch,nguyen2016budgeted}.

\bibliographystyle{IEEEtran}
\bibliography{tnnls2019}

%

\end{document}